\documentclass[journal]{IEEEtran}
\usepackage{amsmath, psfrag, blindtext, amssymb, relsize, textcomp, mathrsfs}
\usepackage[export]{adjustbox}
\newcommand{\subparagraph}{}

\usepackage{subcaption,dirtytalk, graphicx, tikz, pgfplots, setspace}
\usepackage[ruled,vlined]{algorithm2e}

\usepackage{multirow, diagbox, epstopdf, caption, titlesec, etoolbox, pgf}
\usetikzlibrary{arrows}
\usetikzlibrary{positioning}
\usetikzlibrary{shapes.geometric, arrows.meta, decorations.markings}
\tikzstyle{startstop} = [rectangle, rounded corners, minimum width=1.25cm, minimum height=0.75cm,text centered, draw=black, fill=red!20, node distance = 1cm]
\tikzstyle{startstop1} = [rectangle, rounded corners, minimum width=1cm, minimum height=0.5cm,text centered, draw=black, fill=blue!30, node distance = 1cm]
\tikzstyle{startstop2} = [rectangle, rounded corners, minimum width=10.5cm, minimum height=1.5cm,text centered, draw=black, fill=gray!10, node distance = 1.5cm]
\tikzstyle{arrow} = [thick,->,>=stealth]
\makeatletter
\patchcmd{\ttlh@hang}{\parindent\z@}{\parindent\z@\leavevmode}{}{}
\patchcmd{\ttlh@hang}{\noindent}{}{}{}
\makeatother
\ifCLASSINFOpdf
 
\else
  
\fi

\begin{document}
\author{Neema Davis, Gaurav Raina, Krishna Jagannathan \thanks{N. Davis, G. Raina and K. Jagannathan are with the Department
of Electrical Engineering, Indian Institute of Technology Madras, Chennai 600 036, India. 
E-mail: \{ee14d212, gaurav, krishnaj\}@ee.iitm.ac.in}}
\title{A Framework for End-to-End Deep Learning-Based Anomaly Detection in Transportation Networks}

\maketitle
\begin{abstract}
We develop an end-to-end deep learning-based anomaly detection model for temporal data in transportation networks. The proposed EVT-LSTM model is derived from the popular LSTM (Long Short-Term Memory) network and adopts an objective function that is based on fundamental results from EVT (Extreme Value Theory). We compare the EVT-LSTM model with some established statistical, machine learning, and hybrid deep learning baselines. Experiments on seven diverse real-world data sets demonstrate the superior anomaly detection performance of our proposed model over the other models considered in the comparison study. 
\end{abstract}


\IEEEpeerreviewmaketitle

\section{Introduction}
% The traces obtained from GPS enabled taxis enable us to extract meaningful statistics regarding the drivers and passengers. Additionally, this information can be used to monitor adverse and/or malicious events. 
The increasing availability of large-scale traffic data sets provides an opportunity to explore them for knowledge discovery in ITS (Intelligent Transportation Systems). The avenues for exploration are numerous, ranging from uncovering traffic patterns \cite{lippi2010collective}, city dynamics \cite{zheng2011urban}, driving directions \cite{yuan2010t}, discovering hot spots in a city \cite{chang2010context}, finding vacant taxis around a city \cite{phithakkitnukoon2010taxi}, predicting taxi demand \cite{davis2018ataxi}, taxi operation patterns \cite{li2011hunting}, to detecting anomalies \cite{chen2013iboat}, among others. %https://ieeexplore.ieee.org/abstract/document/6450098

Various verticals of ITS have always received active research attention in the past. However, the recent emergence of deep learning techniques and their applicability in transportation systems has resulted in a heightened interest in this area \cite{wang2018enhancing}. Consequently, traditional machine learning models in many applications are now being replaced by deep learning techniques, which is reshaping the landscape of intelligent transport networks. Out of the several applications of ITS, the area of anomaly detection has benefited significantly from the application of deep learning-based techniques \cite{chalapathy2019deep}. Anomaly detection aims to find those patterns which are not normally expected from the data. Typical observations from traffic data demonstrate strong spatio-temporal patterns, showing periodicity and strong correlations between adjacent observations. These patterns may vary depending on the time of the day, day of the week, season, or location. Occasional deviations from these patterns can be termed as abnormal events. Information explored from these anomalous events can provide useful guidelines to urban planners. For instance, abnormal traffic event detection can be utilized to help mitigate congestion, plan driving routes, and reduce the imbalance between taxi demand and supply.

% Abnormal event detection can, for instance, be utilized to help mitigate congestion and diagnose bottlenecks. With the variation of data used, anomaly detection can draw different conclusions and hence, it can find uses in various applications. 

Within the transportation domain, anomaly detection has been applied to abnormal trajectory detection \cite{chen2013iboat}, finding atypical regions \cite{kong2018lotad}, obstacle detection \cite{dairi2018obstacle}, congestion analysis \cite{markou2017use}, and irregularities in taxi passenger demand \cite{wittmann2018event}, among others. Anomaly detection also finds extensive use in a wide range of applications such as fraud detection for credit cards, insurance, or health care, intrusion detection for cyber-security, fault detection in safety-critical systems, and military surveillance for enemy activities \cite{chandola2009anomaly}. 

% Usually prediction works, but  Occasionally,there  are  unusual  traffic  events  (such  as  accidents,  jams,  andsevere  weather)  that  disrupt  the  expected  road  traffic  condi-tions.  Detecting  the  occurrence  of  such  events  in  an  online  andreal-time  manner  is  useful  to  drivers  in  planning  their  routesand in the management of the transportation infrastructure.
% , travel demand exceeds the existing travel capacity in many big cities, which leads to many urban traffic problems and negatively impacts citizens’ standard of living

% https://dl.acm.org/citation.cfm?id=3284725 have some references for diff anomaly detection in transportation
\subsection{Related Literature}
\label{relatedworks}
% In the literature, some solutions of anomalous trajectory detection have already been reported
% One simple yet feasible way to find “abnormal” events isto define a normal range of traffic measurements based onexperience and to use thresholds to identify such special events.However, threshold-based methods are neither reliable noradaptive to changing environment. 
%  The problem of forecasting is closely related to many forms of temporal outlier analysis, since outliers are often defined as deviations from expected values (or forecasts).
Traditionally, anomaly detection has been performed using parametric and non-parametric statistical models, data clustering, rule-based systems, mixture models, and SVMs (Support Vector Machines), among others; for extensive surveys, the interested reader can refer to \cite{chandola2009anomaly} and \cite{hodge2004survey}. These traditional models often fail to capture the complex structures in the data. Additionally, as the volume of the data increases, traditional methods may experience difficulties in finding outliers at such a large scale. Hence, the performance of the aforementioned algorithms in detecting outliers might be sub-optimal for real-world sequences. 

In recent years, deep learning-based anomaly detection algorithms have become increasingly popular, with applications in a diverse set of tasks \cite{chalapathy2019deep}. Unsupervised anomaly detection using deep learning has mainly been hybrid in nature. First, the deep neural network learns the complex patterns of the data. Then, the hidden layer representations from this trained network are used as input to traditional anomaly detection algorithms. There are two popular categories of deep learning-based anomaly detection. The first category consists of methods that analyze the reconstruction errors in an auto-encoder trained over the normal data. A deficiency in the reconstruction of a test point indicates abnormality \cite{malhotra2016lstm}. The second class of methods utilizes either an auto-encoder trained over the normal class to generate a low-dimensional embedding, or a neural network to generate predictions. To identify anomalies, one uses classical methods over the embedding or predictions, such as a parametric distribution assumption \cite{malhotra2016lstm}, an OC-SVM (One Class-SVM) \cite{oza2018one}, etc. 

While the currently popular hybrid deep learning-based anomaly detection techniques have proven to be effective in multiple tasks, these neural networks are not customized for anomaly detection. Since the hybrid models extract features using a neural network and feed it to a separate anomaly detection method, they fail to influence the representational learning in the hidden layers. A more advanced variant of this approach combines the encoding and detection steps using an appropriate objective function, which is used to train a single neural model that performs both procedures \cite{ruff2018deep}. In another related research \cite{golan2018deep}, the authors use geometrical transformations to perform end-to-end deep learning-based anomaly detection using CNNs (Convolutional Neural Networks). In \cite{chalapathy2018anomaly}, an OC-SVM objective is implemented in a feed-forward neural network for deep anomaly detection. 

The primary focus of the aforementioned literature is on anomaly detection in the context of image data sets. The anomaly detection techniques tailored for images need not necessarily perform well with time sequences. Therefore, in this study, we aim to develop an end-to-end anomaly detection using LSTM (Long Short-Term Memory) network \cite{gers1999learning}, which is a neural network designed for sequential data. By gathering insights from EVT (Extreme Value Theory) \cite{siffer2017anomaly}, we design an end-to-end LSTM-based anomaly detection model. To the best of our knowledge, an LSTM-based end-to-end deep anomaly detection model for transportation data has not been explored in the literature. Further, our objective function and network weight updation are based on results from EVT. So far, Extreme Value Theory has not been employed in training a neural network model for performing anomaly detection. These features set our research apart from existing literature\footnote{A part of this work has been presented as a conference paper \cite{davis2019lstm}.}.

% http://papers.nips.cc/paper/8183-deep-anomaly-detection-using-geometric-transformations.pdf
\subsection{Our Contributions}
We propose an end-to-end deep anomaly detection algorithm, and compare the model against several baseline models: (i) parametric GARCH (Generalized Auto Regressive Conditional Heteroskedasticity) model, (ii) non-parametric OC-SVM model, and (iii) hybrid LSTM anomaly detection models based on different detection rules. The detection rules used in hybrid deep anomaly detection model are based on the Gaussian distribution, Tukey's method, and EVT. The key findings obtained by comparing the traditional and deep learning-based models are outlined below.
\begin{enumerate}
    \item This study develops an end-to-end deep anomaly detection algorithm for temporal data based on an LSTM network and an objective function derived from EVT.
    \item Our proposed EVT-LSTM model outperforms several statistical, machine learning, and hybrid deep learning-based algorithms across seven diverse data sets.
    \item We highlight the necessity of a customized neural network model in deep learning-based anomaly detection setting.
    % \item Deep learning-based anomaly detection algorithms outperform statistical and machine learning-based algorithms across seven diverse data sets.
    % \item Proposed end-to-end deep anomaly detection algorithm, EVT-LSTM, exhibits superior detection accuracy when compared to hybrid deep anomaly detection algorithms based on multiple detection rules.
\end{enumerate}
The rest of the paper is organized as follows. In Section \ref{baselinessection}, we explain the traditional baseline models considered for anomaly detection in this study. The hybrid deep anomaly detection model, along with the three detection strategies, is explained in Section \ref{hybridsection}. It is followed by Section \ref{evtlstmsection}, where we introduce our proposed EVT-LSTM model. The experimental settings are provided in Section \ref{experimentalsettingssection}, and the results are outlined in Section \ref{resultssection}. We conclude our work in Section \ref{conclusionsection}. 

\section{Traditional Anomaly Detection} \label{baselinessection}
In this section, we provide brief descriptions of two traditional anomaly detection models considered as baselines in our comparison study. 

\subsection{GARCH Model}
% Anomaly detection using parametric models is based on the principle that any observation that is not generated by the stochastic model assumed is a likely anomaly.  In statistical anomaly detection, parametric techniques are quite popular and 
Parametric statistical models \cite{fox1972outliers} represent one of the early works on outlier detection in time series. Several models were subsequently proposed in the literature for parametric anomaly detection, including ARMA (Auto Regressive Moving Average), ARIMA (Auto Regressive Integrated Moving Average), and EWMA (Exponentially Weighted Moving Average), to list a few \cite{chandola2009anomaly}. We assume that the normal data instances are located at the high probability regions of a stochastic model compared to the anomalies that have a low probability. A common practice followed here is to either assume a distribution for the anomalies \cite{eskin2000anomaly} or fit a regression model to the data \cite{chen2005simultaneous}. 

A regression-based anomaly detection technique involves two steps: (a) the regression model is used to model the data, (b) the residuals, \emph{i.e.}, the part not explained by the regression model, are used to determine the anomaly scores. 
% Often, the magnitude of the residuals can be used as the anomaly score for the test data. 
A popular choice for regression-based anomaly detection is the GARCH model \cite{engle2001garch}, which is often applied to financial time-series. A GARCH process is often preferred over other regression models such as ARMA because it imposes a specific structure on the conditional variance of the process. The variance is not assumed to be a constant, making the series non-stationary in nature and rendering them suitable for real-world scenarios. Essentially, the GARCH process models the error variance of the time-series as an ARMA process. The AR part models the variance of the residuals and the MA portion models the variance of the process. The time series {$\epsilon_t$} at each instance $t$ is given by:
\begin{equation}
    \epsilon_t = \sigma_t w_t,
\end{equation}
where, {$w_t$} is discrete white noise with zero mean and unit variance, and ${\sigma_t}^2$ is given by:
\begin{equation}
      {\sigma_t}^2 = \delta_0 + \sum_{i=1}^{r} \delta_i {\sigma_{t-i}}^2 + \sum_{i=1}^{s} \gamma_i {\epsilon_{t-i}}^2,  
\end{equation}
where, $\delta_i$ and $\gamma_i$ are the parameters of the model. In other words, {$\epsilon_t$} is a Generalized Auto Regressive Conditional Heteroskedastic model of order r and s, denoted by GARCH(r, s).

Parametric methods allow the model to be evaluated very rapidly for new instances and are suitable for large data sets; the model grows only with model complexity and not the data size. However, they limit their applicability by enforcing a predetermined distribution to the data. These approaches are accurate only if the data fits the chosen distribution model. The non-parametric approach described below can overcome this disadvantage associated with parametric models.

\subsection{OC-SVM Model}
Non-parametric methods such as SVMs \cite{scholkopf2002learning} apply local kernel models rather than a single global distribution model to the data. Their popularity stems from the ability to combine speed and low complexity growth of parametric methods with the model flexibility of non-parametric methods. Kernel-based methods estimate the density distribution of the input space and identify outliers as lying in regions of low density.

Typically, the SVM model is given a set of training examples labeled as belonging to one of two classes. The model tries to divide the training sample points into two categories by creating a boundary while penalizing training samples that fall on the wrong side of the boundary. The SVM model can then make predictions by assigning points to either side of the boundary. For anomaly detection applications, the training examples are often limited. Therefore, SVMs are more popularly applied in a one-class setting here, where the SVM model is trained on data that has only one class, that is the \emph{normal} class. This is particularly useful in anomaly detection because by inferring the properties of the normal class, the examples that deviate from the normal class can be identified. The SVM model needs a kernel function that can map the original non-linear observations into a higher-dimensional space in which they are separable. Commonly used kernel functions are linear, sigmoid, Gaussian, and RBF (Radial Basis Function) \cite[Chapter 2]{scholkopf2002learning}. During the testing phase, if a test instance falls within the learned region, it is declared as normal, else it is deemed as anomalous. 

The SVM model requires a kernel function, which has to be carefully tuned for obtaining good classification accuracy. Further, the anomaly detection is supervised in nature; it requires prior knowledge of the labels. On the other hand, the recently developed anomaly detection models based on neural networks can perform unsupervised anomaly detection, and hence, has seen widespread use over the SVM model for anomaly detection lately.

\section{Hybrid Deep Anomaly Detection} \label{hybridsection}
The suitability of neural network models for anomaly detection originates from their unsupervised learning nature and the ability to learn highly complex non-linear sequences.  When presented with normal non-anomalous data, the neural network can learn and capture the normal behavior of the system. Later, when the model encounters a data instance that deviates significantly from the rest of the set, it generates a high prediction error, suggesting at anomalous behavior. This form of prediction-based anomaly detection is hybrid in nature as it requires the application of a set of detection rules on the errors obtained from the network. Often, the decision rules employed are traditional statistical or machine learning anomaly detection algorithms. Popular detection techniques involve thresholding the prediction errors \cite{wang2011statistical}, assuming an underlying parametric distribution on the prediction errors \cite{malhotra2016lstm}, or applying machine learning techniques such as an SVM model on the errors \cite{ergen2017unsupervised}. We now briefly describe the prediction model and detection rules considered in our study. 

\subsection{Prediction Model}
We use the LSTM network as the time-series prediction model. They are state of the art neural network models which are widely used in sequence learning applications \cite{gers1999learning}. We feed the recent $l_b$ number of values of every data set into the model, and the model outputs $l_a$ number of forecasts. The $l_b$ and $l_a$ are known as look-back and look-ahead respectively. Dropout and early stopping are employed to avoid over-fitting.

Each data set is divided into a training set, a validation set, and a test set. The model learns from the training data and validates its performance on the hold-out validation data. The training set is assumed to be free of anomalies. This is a reasonable assumption in real-world scenarios where instances of normal behavior may be available in abundance, but instances of anomalous behavior are rare. The validation and test set are mixtures of anomalous and non-anomalous data instances. The prediction model is trained on normal data without any anomalies, \emph{i.e.}, on the training data, so that it learns the normal behavior of the time-series. Once the model is trained, anomaly detection is performed on the test set, by using the prediction errors as anomaly indicators. In this paper, the prediction error is defined as the absolute difference between the input received at time $t$ and its corresponding prediction from the model at $t$. 

We consider three detection techniques by which the prediction errors can be used to set an anomaly threshold: (i) the Gaussian-based detection rule that makes assumptions about the parent distribution, (ii) the Tukey's method based detection rule that does not make any assumptions on the distribution, and (iii) the EVT-based detection rule that makes assumption about the tail of the distribution, but not about the parent distribution. If any prediction error value lies outside of the chosen threshold, then the corresponding input can be considered as a possible anomaly. The detection rules considered are as follows:

\subsection{Gaussian-based Detection \cite{malhotra2016lstm}} 
One of the earliest and popular works in prediction-based anomaly detection setting \cite{malhotra2016lstm} assumes that the prediction errors from the training set follow a Gaussian distribution. The prediction errors obtained from the LSTM model is fit to a Gaussian distribution. The mean, $\mu$, and variance, $\sigma^2$, of the Gaussian distribution are computed using MLE (Maximum Likelihood Estimation) \cite{myung2003tutorial}. The Log PDs (Probability Densities) of errors are calculated based on the parameters estimated and used as anomaly scores. A low value of Log PD indicates that the likelihood of an observation being an anomaly is high. A validation set containing both normal data and anomalies is used to set a threshold $\tau_{g}$ on the Log PD values. The threshold is chosen such that it can separate all the anomalies from normal observations while incurring as few false positives as possible. The threshold is then evaluated on a separate test set.
\subsection{Tukey's Method Based Detection \cite{wang2011statistical}} 
Tukey's method uses quartiles to define an anomaly threshold. It makes no distributional assumptions and does not depend on the knowledge of a mean or a standard deviation. In Tukey's method, a possible outlier lies outside the threshold $\tau_{t} = Q_3 + 3 \times (Q_3 - Q_1)$, where $Q_1$ is the lower quartile or the  $25^{th}$ percentile, and $Q_3$ is the upper quartile or the $75^{th}$ percentile. The metric $Q_3 - Q_1$ is known as the interquartile distance. The prediction errors obtained from the training, validation,  and test sets are concatenated, and the lower quartiles and interquartile distances are calculated. The values lying outside $\tau_{t}$ are identified as potential outliers. 
\subsection{EVT-based Detection \cite{siffer2017anomaly}}
Let $X$ be a random variable and $F(x) = P(X \leq x)$ be its CDF (Cumulative Distribution Function). The tail of the distribution is given by $\Tilde{F}(x) = P(X > x)$. The probability $P(X > x)$ tends to zero for the extreme events in the system. A key result from EVT \cite{de2007extreme} suggests that the distribution of the extreme values is not highly sensitive to the parent data distribution. This result enables us to accurately compute probabilities without first estimating the underlying distribution. Under a weak condition, the extreme events have the same kind of distribution, regardless of the parent distributions, known as the EVD (Extreme Value Distribution):
    \begin{equation}
         G_{(\sigma,\ \gamma)} : y \rightarrow \text{exp} \Big ( -\Big(1+\gamma \frac{y}{\sigma} \Big)^{-\frac{1}{\gamma}} \Big),\ \gamma \in \mathbb{R},\ 1 + \gamma \frac{y}{\sigma} > 0,
    \end{equation}
where, $\sigma$ is the scale parameter, and $\gamma$ is the extreme value index of the distribution. Based on the value $\gamma$ takes, the tail distribution can be Fr\'echet ($\gamma\ >\ 0$), Gumbel ($\gamma\ =\ 0$), or Weibull ($\gamma\ <\ 0$).  By fitting an EVD to the unknown input distribution tail, it is then possible to evaluate the probability of potential extreme events. In some recent work \cite{siffer2017anomaly}, the authors use results from EVT to detect anomalies in a uni-variate data stream, following the POTs (Peaks-Over-Thresholds) approach. Based on an initial threshold $T$, the POTs approach attempts to fit a GPD (Generalized Pareto Distribution) to the excesses, $X-T$. In other words, rather than fitting an EVD to the extreme values of $X$, the POTs approach fits a GPD to the excesses $X-T$. To compute the maximum likelihood estimates for GPD, we follow the procedure outlined by \cite{grimshaw1993computing}. Once the parameters are obtained, the threshold $\tau_e$ can be computed as:
\begin{equation}
        \tau_e = T + \frac{\hat{\sigma}}{\hat{\gamma}} \Big( \Big(\frac{qn}{N_t} \Big)^{-\hat{\gamma}} -1 \Big ), \label{evtthreshold}
\end{equation}
where, $\hat{\sigma}$  and $\hat{\gamma}$ are the estimated parameters of the GPD, $q$ is some desired probability, $n$ is the total number of observations, and $N_t$ is the number of peaks, \emph{i.e.}, the number of $X_i$ s.t. $X_i > T$. The probability $P(X > \tau_e)$ is calculated for all the observations and those data instances with $P(X > \tau_e) < q$ can be considered as plausible anomalies. The authors in \cite{siffer2017anomaly} recommend choosing a value for $q$ within [$10^{-3}$, $10^{-5}$] and $T$ as the 98\% quantile, which we follow in our study. More details of this algorithm can be found in \cite{siffer2017anomaly}. 

\section{End-to-End Deep Anomaly Detection} \label{evtlstmsection}
In Section~\ref{relatedworks}, we highlighted the need for developing end-to-end deep learning-based anomaly detection models, especially for temporal data. An end-to-end deep anomaly detection technique involves modifying the objective function of a deep learning model such as an LSTM or a CNN. Modifications are introduced so that the models that were formerly learning patterns for forecasting will now learn to detect deviations from the normal behavior. Instead of first predicting using a neural network and then feeding the predictions to a separate post-processing technique, the outputs of an end-to-end deep anomaly detection model can be directly interpreted as anomaly scores. In \cite{ruff2018deep}, the authors combine a CNN with an SVDD (Support Vector Deep Description) objective. The SVDD is a technique similar to the OC-SVM, where a hyper-sphere is used to separate the data instead of a hyper-plane. 

Let $\phi(\cdot ; \mathcal{W}): \mathcal{X}\rightarrow\mathcal{Y}$ be a neural network with $L$ layers and a set of weights $\mathcal{W} = \{\mathbf{W}^1, \ldots, \mathbf{W}^L\}$. This network maps data from an input space $\mathcal{X} \subseteq \mathbb{R}^p$ to an output space $\mathcal{Y} \subseteq \mathbb{R}^q$. That is, $\phi(\mathbf{x}; \mathcal{W}) \in \mathcal{Y}$ is the network representation of $\mathbf{x} \in \mathcal{X}$ given by the network $\phi$ with parameters $\mathcal{W}$. The One-Class Deep SVDD objective given in \cite{ruff2018deep}, for a CNN model with input \{$\mathbf{x}_1, \ldots, \mathbf{x}_N$\}, is as follows:
\begin{equation}
    \underset{\mathcal{W}}{\text{min}}\ \frac{1}{N} \sum_{i=1}^N || \phi(\mathbf{x}_i; \mathcal{W}) - \mathbf{c} ||^2 + \frac{\lambda}{2} \sum_{l=1}^L {|| \mathbf{W}^l ||_F}^2. \label{svdd}
\end{equation}
The first term in the quadratic loss objective function penalizes the distance between every network representation $\phi(\mathbf{x}_i; \mathcal{W})$ and the center of the hyper-sphere $\mathbf{c}$. The second term penalizes the network weights by employing a network weight decay regularizer with hyper-parameter $\lambda > 0$, where $||.||_F$ denotes the Frobenius norm. In \cite{ruff2018deep}, the $\mathbf{c}$ was fixed as the mean of the network predictions that results from performing an initial forward pass on the training data samples. The experiments were conducted for MNIST and CIFAR-10 image data sets. 

In order to develop a similar model for time-sequences, we implement the aforementioned objective function in an LSTM model. Interestingly, we find that while this quadratic loss objective function works satisfactorily for anomaly detection in images, it does not fare well for temporal data. When adopted in the LSTM network, we notice that Eqn.~\ref{svdd} minimizes the distance between the predictions and their initial mean by reducing the magnitude of the predictions, resulting in a large fraction of false positives. This behavior suggests that an objective function that directly minimizes the network predictions might not be a sensible choice for anomaly detection in temporal data. We recall that the success of hybrid deep learning-based anomaly detection algorithms was mainly attributed to an efficient threshold based on the prediction errors. Therefore, it is natural to explore an objective function that minimizes the prediction errors and not the actual predictions. 

Further, in our recent work \cite{davis2019lstm}, after comparing different detection strategies for hybrid deep anomaly detection, we noticed the potential of a strategy based on extreme values. We found that an EVT-based detection rule performed better than other popular detection techniques. The superior performance of an EVT-based strategy in a deep learning setting encouraged us to integrate EVT into the objective function of the LSTM model, leading to an end-to-end deep anomaly detection model. 

\subsection{EVT-LSTM model}
In our study, the inputs $\{\mathbf{x}_1, \ldots, \mathbf{x}_N\}$ in $\mathcal{X} \subseteq \mathbb{R}^p$ are mapped to the set $\{{y}_1, \ldots, {y}_N\}$ in $\mathcal{Y} \subseteq \mathbb{R}$. Our EVT-LSTM model is based on the objective function given as follows:
\begin{equation}
    \underset{\mathcal{W}}{\text{min}}\ \frac{1}{N} \sum_{i=1}^N || \text{E}(\phi(\mathbf{x}_i; \mathcal{W})) - {\tau_e} ||^2 + \frac{\lambda}{2} \sum_{l=1}^L {|| \mathbf{W}^l ||_F}^2. \label{evtlstmobj}
\end{equation}
Here, instead of minimizing the distance between the network representations and the mean obtained after an initial forward pass as in Eqn.~\ref{svdd}, we minimize the Euclidean distance between every absolute prediction error $\text{E}(\phi(\mathbf{x}_i; \mathcal{W}))$ and a threshold ${\tau_e}$. The threshold ${\tau_e}$ is obtained from Eqn.~\ref{evtthreshold}, and is updated periodically during the training phase. This form of optimization is called an alternating minimization approach and has been used with similar objective functions in related literature \cite{ruff2018deep, chalapathy2018anomaly}. The objective functions in these related literature minimized a function of the predictions obtained from image data sets. On the other hand, our objective function (Eqn.~\ref{evtlstmobj}) optimizes a function of the prediction errors. Our proposed algorithm is given in Algorithm~\ref{evtlstm}.
\begin{algorithm}
\SetAlgoLined
\setstretch{1}
    \textbf{Input:} Set of examples $(\mathbf{x}_{n:}, y_{n:}),\ n \in  \{1, \ldots, N\}$ \\
    %Set of examples $\{\mathbf{x}_1, \ldots, \mathbf{x}_N\}, \{{y}_1, \ldots, {y}_N\} \in \mathcal{Y}$\\
    \textbf{Output:} Set of decision scores $s_{n:},\ n \in \{1, \ldots, N\}$\\
    %Set of decision scores $\mathcal{S} = \{s_1, \ldots, s_N\}$\\
    \textbf{Initialization:} Threshold ${\tau_e} \leftarrow {0}$ \\
     \While{convergence criteria unmet} {
       Update weights of the network using Eqn.~\ref{evtlstmobj} \\
        \For{once in every $k$ epochs} { 
        Calculate prediction errors, ${\text{E}}(\hat{y}_{n:}) = |\hat{y}_{n:}-y_{n:}|$ \\
        ${T} \leftarrow$ InitThreshold(${E}(\hat{y}_{n:})$) \\
        Excesses $\leftarrow \{{\text{E}}(\hat{y}_{n:}) -{T} | \text{E}(\hat{y}_{n:}) > {T}\}$ \\
        Fit a GPD to excesses by using MLE and find $\hat{\gamma}$, $\hat{\sigma}$ \\
        Update ${\tau_e}$ using Eqn.~\ref{evtthreshold}\\
        }
        }
    Compute decision score $s_{n} = |\hat{y}_{n}-y_{n}|-\tau_e$ for each $\mathbf{x}_n$\\
    \eIf{$s_n \geq 0$}{
       $\mathbf{x}_n$ is anomalous \\
   }{
    $\mathbf{x}_n$ is non-anomalous
  }
   \caption{The training process of the proposed EVT-LSTM model. The threshold ${\tau_e}$ is updated every $k$ = 20 epochs. }\label{evtlstm}
\end{algorithm}

The threshold ${\tau_e}$ is initialized to zero at the beginning of the experiment. During the training phase, the LSTM model tries to optimize the objective function given in Eqn.~\ref{evtlstmobj}. The prediction errors on the training set are calculated every $k$ epochs. The 98\% empirical quantile of the errors is chosen to set an initial threshold ${T}$ in InitThreshold($\text{E}(\phi(\mathbf{x}; \mathcal{W}))$). The excesses occurring above ${T}$ are fit to a GPD using MLE, and the parameters $\hat{\gamma}$ and $\hat{\sigma}$ are estimated. Then, using Eqn.~\ref{evtthreshold}, we calculate the new value for the threshold ${\tau_e}$. The objective function (Eqn.~\ref{evtlstmobj}) is updated with this recent value of threshold obtained. The next $k$ epochs use the modified objective function to train the model, after which the threshold ${\tau_e}$ is again calculated and updated. The training stops when either the convergence is achieved, or the maximum number of epochs is reached. Finally, on a test set, the decision scores are calculated and used for classifying instances as anomalous or non-anomalous. 

\section{Experimental Settings} \label{experimentalsettingssection}
In this section, we discuss the data sets considered, evaluation metrics used, and the procedure for choosing parameters for each anomaly detection model. 

\subsection{Description of Data Sets}
We consider seven real-world data sets in our comparison study: three road traffic-based data sets, two taxi demand data sets, and two data sets from other application domains. The travel time, vehicle occupancy, and traffic speed data sets considered are real-time data, obtained from a traffic detector and collected by the Minnesota Department of Transportation. Discussions on these traffic data sets are available at the Numenta Anomaly Benchmark GitHub repository\footnote{https://github.com/numenta/NAB/tree/master/data}. The NYC (New York City) taxi demand data set is publicly available at \cite{nyc} and contains the trip details of government-run street hailing taxis. The Bengaluru taxi demand data set is acquired from a leading private Indian transportation company dealing with app-based taxi rental services. The ECG (electrocardiogram) data is obtained from \cite{keogh2005hot} and has annotations from a cardiologist to indicate the unusual heartbeat patterns. Bitcoin historic prices are obtained from coindeskr\footnote{https://cran.r-project.org/package=coindeskr} package, R.     

Brief descriptions of the data sets used are given below. 
\begin{enumerate}
    \item Vehicular Travel Time: The data set is obtained from a traffic sensor and has 2500 readings from July 10, 2015, to September 17, 2015, with eight marked anomalies.  
    \item Vehicular Speed: The data set contains the average speed of all vehicles passing through the traffic detector. A total of 1128 readings for the period September 8, 2015 - September 17, 2015, is available. There are three marked unusual sub-sequences in the data set.
    \item Vehicle Occupancy: There are a total of 2382 readings indicating the percentage of the time, during a 30-second period, that the detector sensed a vehicle. The data is available for a period of 17 days, from September 1, 2015, to September 17, 2015, and has two marked anomalies. 
    \item NYC (New York City) Taxi Demand \cite{nyc}: The publicly available NYC data set contains the pick-up locations and time stamps of street hailing yellow taxi services from the period of January 1, 2016, to February 29, 2016. We pick three time-sequences (S1, S2, and S3) with clearly apparent anomalies from data aggregated over 15 minute time periods in 1 $\text{km}^2$ grids. 
    \item Bengaluru Taxi Demand: This data set has GPS traces of passengers booking a taxi by logging into the service provider's mobile application. Similar to the NYC data set, this data is also available for January and February 2016. We aggregate the data over 15 minute periods in 1 $\text{km}^2$ grids and pick three sequences with clearly visible anomalies.
    \item ECG (Electrocardiogram) \cite{keogh2005hot}: There are a total of 18000 readings, with three unusual sub-sequences labeled as anomalies. The data set has a repeating pattern, with some variability in the period length.
    \item Bitcoin Prices: Historical bitcoin prices are available for the period from January 1, 2017, to May 27, 2019. The fraction of anomalies in this data set of 877 readings are observed to be 0.06\%, most of them occurring around the beginning of the year 2018. 
\end{enumerate}

\subsection{Evaluation Metrics}
We consider three evaluation metrics for comparing our models: (i) Precision, $P$, (ii) Recall, $R$, and (iii) F1-score, $F1$, which is the harmonic mean of Precision and Recall. Min-max normalization is performed on every data set before modeling and evaluation.

\begin{enumerate}
    \item Precision, $P$:
    \begin{equation}
      P = \frac{\text{True positives}}{\text{True positives + False positives}},
    \end{equation}
    \item Recall, $R$:
    \begin{equation}
        R = \frac{\text{True positives}}{\text{True positives + False negatives}},
    \end{equation}
    \item F1-score, $F1$:
    \begin{equation}
        F =  2 \cdot \frac{P \times R}{P + R}.
    \end{equation}
\end{enumerate}
True positives are the anomalous instances that have been correctly classified as anomalies by the model. Similarly, true negatives are the instances correctly identified as non-anomalous data. False positives are the non-anomalies incorrectly classified as anomalous, and false negatives are the incorrectly identified anomalies. Since F1-score summarizes both Precision and Recall, we consider the model with the highest F1-score as the superior anomaly detection technique. 
\subsection{Parameter Selection}
In order to perform efficient anomaly detection, it is necessary to set appropriate hyper-parameters and anomaly thresholds for each model. The suitable set of parameters and thresholds vary with the use case considered. Below, we briefly discuss the procedures through which the parameters are shortlisted for each anomaly detection model.
\begin{table*}[]
\centering
\scalebox{1}{
{\renewcommand{\arraystretch}{1.1}
\begin{tabular}{lccc}
\hline
\multicolumn{2}{l}{\textbf{Data Sets}}                                                         & \textbf{Model}                                                             & \textbf{Threshold} \\ \hline \hline
\multicolumn{2}{ l  }{\begin{tabular}[c]{@{}l@{}}Vehicular Travel Time\end{tabular}} & \begin{tabular}[c]{@{}c@{}}ARIMA(1, 0, 3)-GARCH(1, 1)\end{tabular} & 0.016                    \\ \hline
\multicolumn{2}{ l  }{Vehicular Speed}                                                  & \begin{tabular}[c]{@{}c@{}}ARIMA(0, 1, 4)-GARCH(1, 1)\end{tabular} & 0.036                    \\ \hline
\multicolumn{2}{ l  }{Vehicle Occupancy}                                                & \begin{tabular}[c]{@{}c@{}}ARIMA(0, 1, 1)-GARCH(1, 1)\end{tabular} & 0.433                        \\ \hline
\multirow{3}{*}{\begin{tabular}[c]{@{}l@{}}NYC Taxi \\ Demand\end{tabular}}       & S1 & \begin{tabular}[c]{@{}c@{}}ARIMA(0, 1, 3)-GARCH(1, 1)\end{tabular} & 0.009                    \\ %\cline{2-4} 
                                                                                  & S2 & \begin{tabular}[c]{@{}c@{}}ARIMA(3, 0, 4)-GARCH(1, 1)\end{tabular} & 0.047                    \\  
                                                                                  & S3 & \begin{tabular}[c]{@{}c@{}}ARIMA(2, 1, 2)-GARCH(2, 2)\end{tabular} & 0.051                    \\ \hline
\multirow{3}{*}{\begin{tabular}[c]{@{}l@{}}Bengaluru Taxi\\  Demand\end{tabular}} & S1 & \begin{tabular}[c]{@{}c@{}}ARIMA(1, 0, 3)-GARCH(1, 2)\end{tabular} & 0.064                    \\ 
                                                                                  & S2 & \begin{tabular}[c]{@{}c@{}}ARIMA(3, 1, 3)-GARCH(1, 1)\end{tabular} & 0.003                    \\ 
                                                                                  & S3 & \begin{tabular}[c]{@{}c@{}}ARIMA(1, 0, 1)-GARCH(1, 2)\end{tabular} & 0.060                    \\ \hline
\multicolumn{2}{ l  }{Electrocardiogram}                                                & \begin{tabular}[c]{@{}c@{}}ARIMA(4, 1, 2)-GARCH(1, 1)\end{tabular} & $10^{-6}$                 \\ \hline
\multicolumn{2}{ l  }{Bitcoin Prices}                                                   & \begin{tabular}[c]{@{}c@{}}ARIMA(2, 1, 2)-GARCH(1, 1)\end{tabular} & 0.025                    \\ \hline
\end{tabular}}
}
\caption{Appropriate ARIMA($p$, $d$, $q$)-GARCH($r$, $s$) models obtained for each data set, by varying $p$, $q$ in the range [0, 5], $d$ in [0, 1], and $r$, $s$ in [1, 2]. The anomaly thresholds are obtained from a hold-out validation set, so that as few false positives are incurred.}
\label{arima-garch}
\end{table*}

\subsubsection{GARCH Model}
% \emph{For anomaly detection using GARCH model, we follow the steps given below:
% \begin{itemize}
%     \item Model the time-series using appropriate ARIMA model.
%     \item Use correlogram to check whether the residuals of this model fit possess evidence of Conditional Heteroskedastic behaviour.
%     \item Fit suitable GARCH model to the residuals.
%     \item Obtain anomaly scores based on the deviation of GARCH predictions from the actual values.
%     \item Set an anomaly threshold based on the validation set and test it on a test set.
% \end{itemize}}
For every data set, time-sequences are generated based on the training data. For Bengaluru and NYC taxi demand data sets, the temporal aggregation is performed at sampling periods of 15 minutes.  Then, by varying the $p$, $q$, and $d$ parameters of an ARIMA($p$, $d$, $q$) process between [1, 5], appropriate models are chosen for every time-sequence. The residuals obtained from fitting the ARIMA processes are then modeled as suitable GARCH($r$, $s$) processes. We find that suitable values for parameters $r$ and $s$ often lie in the range [1, 2]. Once appropriate models are developed, anomaly scores are obtained based on the deviation of the GARCH predictions from the actual values. An anomaly threshold is set based on the validation set and examined on a test set. The parameters of the fitted ARIMA-GARCH models, along with the anomaly thresholds are given in Table \ref{arima-garch}.

\subsubsection{OC-SVM Model}
% \textit{For anomaly detection using OC-SVM model, we follow the steps given below:
% \begin{itemize}
%     \item The OC-SVM runs on a training set ?
%     \item Based on the performance of the OC-SVM on a holdout validation set, the parameters are chosen.
%     \item OC-SVM with the chosen parameters is tested on a test set to detect anomalies
% \end{itemize}}
Appropriate kernel functions are crucial for satisfactory anomaly detection performance of SVMs, and the choices vary with the data sets considered. In our study, we consider Linear, RBF, Polynomial, and Sigmoid kernels. Another important parameter is the kernel coefficient $\alpha$ for the RBF, Polynomial, and Sigmoid kernels. After varying $\alpha$ in the range [0.0001, 0.1], a value of 0.0001 is found to suit most of the data sets considered. For every use case, multiple SVM models ran on the training data, with different parameters chosen from the range of values considered. Then, suitable choices are made by observing the classification accuracy on a hold-out validation set. Finally, the best OC-SVM model obtained is used to detect anomalies on a test set. The shortlisted OC-SVM models are given in Table \ref{oc-svm}.
\begin{table}[]
\centering
\scalebox{0.95}{
{\renewcommand{\arraystretch}{1.1}
\begin{tabular}{ l c  l }
\hline
\multicolumn{2}{ l  }{\textbf{Data Sets}}                                                         & \textbf{Kernel Setting}                             \\ \hline \hline
\multicolumn{2}{ l  }{Vehicular Travel Time}                                            & RBF(0.0001)                   \\ \hline
\multicolumn{2}{ l  }{Vehicular Speed}                                                  & Poly(0.0001)                 \\ \hline
\multicolumn{2}{ l  }{Vehicle Occupancy}                                                & RBF(0.0001)                  \\ \hline
\multirow{3}{*}{\begin{tabular}[c]{@{}l@{}}NYC Taxi \\ Demand\end{tabular}}       & S1 & \multirow{3}{*}{RBF(0.0001)} \\ %\cline{2-2}
                                                                                  & S2 &                                    \\ 
                                                                                  & S3 &                                    \\ \hline
\multirow{3}{*}{\begin{tabular}[c]{@{}l@{}}Bengaluru Taxi \\ Demand\end{tabular}} & S1 & \multirow{3}{*}{RBF(0.0001)} \\ %\cline{2-2}
                                                                                  & S2 &                                    \\ %\cline{2-3} 
                                                                                  & S3 &                   \\ \hline
\multicolumn{2}{ l  }{Electrocardiogram}                                                & Linear              \\ \hline
\multicolumn{2}{ l  }{Bitcoin Prices}                                                   & Sigmoid(0.0001)              \\ \hline
\end{tabular}}}
\caption{The shortlisted OC-SVM models for the data sets considered. We consider Linear, Sigmoid, Polynomial, and RBF kernels, and vary $\alpha$ between [0.0001, 0.1].}
\label{oc-svm}
\end{table}

\begin{table*}[htb!]
\centering
\scalebox{0.9}{
{\renewcommand{\arraystretch}{1.1}
\begin{tabular}{ l  l  c }
\hline
\multicolumn{1}{ c  }{\textbf{Data Sets}} & \multicolumn{1}{c  }{\textbf{LSTM Architecture}}    \\ \hline \hline
Vehicular Travel Time       & \begin{tabular}[l]{@{}l@{}}1 Recurrent layer: \{20\}, Dropout: 0.2,\\ 1 Dense layer: \{1\}, Learning rate: 0.01\end{tabular}        \\ \hline
Vehicular Speed                & \begin{tabular}[l]{@{}l@{}}1 Recurrent layer: \{60\}, Dropout: 0.19,\\ 1 Dense layer: \{1\}, Learning rate: 0.0001\end{tabular}      \\ \hline
Vehicle Occupancy         & \begin{tabular}[l]{@{}l@{}}1 Recurrent layer: \{50\}, Dropout: 0.23,\\ 1 Dense layer: \{1\}, Learning rate: 0.0001\end{tabular}       \\ \hline
NYC Taxi Demand                & \begin{tabular}[l]{@{}l@{}}2 Recurrent layers: \{50, 20\}, Dropout: 0.4,\\ 1 Dense layer:\{24\}, Learning rate: 0.0001\end{tabular}  \\ \hline
Bengaluru Taxi Demand          & \begin{tabular}[l]{@{}l@{}}2 Recurrent layers: \{20, 10\}, Dropout: 0.25,\\ 1 Dense layer:\{24\}, Learning rate: 0.0001\end{tabular}  \\ \hline
Electrocardiogram                        & \begin{tabular}[l]{@{}l@{}}2 Recurrent layers: \{60, 30\}, Dropout: 0.1,\\ 1 Dense layer:\{5\}, Learning rate: 0.05\end{tabular}        \\ \hline
Bitcoin Prices            & \begin{tabular}[l]{@{}l@{}}1 Recurrent layer: \{10\}, Dropout: 0.1,\\ 1 Dense layer: \{1\}, Learning rate: 0.0001\end{tabular}      \\ \hline
\end{tabular}}}
\caption{The LSTM architectures for the data sets considered. The optimal set of hyper-parameters for each data set is chosen after running the TPE (Tree-structured Parzen Estimator) Bayesian Optimization algorithm.  }
\label{settings}
\end{table*}
\subsubsection{Hybrid LSTM Models}
For a neural network model, hyper-parameters define the high-level features of the model, such as its complexity, or capacity to learn. The important hyper-parameters include the number of hidden recurrent layers, dropout values, learning rate, and the number of units in each layer. We use the TPE (Tree-structured Parzen Estimator) Bayesian Optimization \cite{bergstra2011algorithms} to select these hyper-parameters. The output layer is a fully connected dense layer with linear activation. The Adam optimizer \cite{kingma2014adam} is used to minimize the Mean Squared Error objective function. All LSTM-based models ran for 100 epochs with a batch size of 64.

The chosen set of parameters for each data set is given in Table \ref{settings}. We follow the same model settings as \cite{singh2017anomaly} for the ECG data set. For the traffic speed, travel time, vehicle occupancy, and bitcoin prices data sets, the limited availability of readings suggested look-back and look-ahead times of 1 each. We have over 10 million points for the New York and Bengaluru cities, allowing for a large look-back time. The considerable amount of data in these two cases allows the LSTM to learn better representations of the input data, aiding the anomaly detection process.

The false positive regulators are the parameters that impact the performance of the detection algorithms. The false positive regulator for the Gaussian-based detection rule, $\tau_g$, is chosen for each time-sequence such that the F1-score on the validation errors is maximized. The thresholds, $\tau_t$, for Tukey's method are directly obtained from the entire set of prediction errors, based on a simple quantile calculation. For both hybrid and end-to-end EVT-LSTM deep learning models, we follow similar procedures to set the parameters for EVT rule. As mentioned earlier, an initial threshold $T$ has to be chosen for the EVT-based detection, typically 98\% quantile. The false positive regulator for the EVT-based anomaly detection, $q$, is set from an initialization data stream. We set $q$ using the same initialization stream that is used for setting $T$. The initialization stream contains the prediction errors from the training and validation sets. The probability $q$ is chosen so that the EVT-based anomaly detection picks up all the anomalies from the initialization stream. The chosen values for the false positive regulators of the hybrid LSTM-based techniques are given in Table \ref{lstm-based}.

\begin{table*}[]
\begin{center}
{\renewcommand{\arraystretch}{1.1}
\begin{tabular}{ l c  c c c }
\hline
\multicolumn{2}{ l  }{\multirow{2}{*}{\textbf{Data Sets}}}                                        & \multicolumn{3}{c }{\textbf{Hybrid LSTM Models}} \\ \cline{3-5} 
\multicolumn{2}{ l  }{}                                                                 & Gaussian ($\tau_g$)      & Tukey ($\tau_t$)       & EVT ($q$)      \\ \hline \hline
\multicolumn{2}{ l  }{Vehicular Travel Time}                                            & -20           & 572.9       & $10^{-4}$     \\ \hline
\multicolumn{2}{ l  }{Vehicular Speed}                                                  & -18           & 24.4        & $10^{-3}$     \\ \hline
\multicolumn{2}{ l  }{Vehicle Occupancy}                                                & -23           & 12.9        & $10^{-5}$     \\ \hline
\multirow{3}{*}{\begin{tabular}[c]{@{}l@{}}NYC Taxi \\ Demand\end{tabular}}       & S1 & -19           & 12.1        & $10^{-5}$ \\ %\cline{2-5} 
                                                                                  & S2 & -17           & 12.8        & $10^{-5}$     \\  
                                                                                  & S3 & -15           & 10.5        & $10^{-5}$     \\ \hline
\multirow{3}{*}{\begin{tabular}[c]{@{}l@{}}Bengaluru Taxi \\ Demand\end{tabular}} & S1 & -25           & 33.5        & $10^{-4}$     \\  
                                                                                  & S2 & -18           & 27.1        & $10^{-4}$     \\  
                                                                                  & S3 & -25           & 14.0        & $10^{-4}$  \\ \hline
\multicolumn{2}{ l  }{Electrocardiogram}                                                & -23           & 0.1        & $10^{-4}$     \\ \hline
\multicolumn{2}{ l  }{Bitcoin Prices}                                                   & -17           & 12961.8     & $10^{-3}$     \\ \hline
\end{tabular}}
\end{center}
\caption{The chosen false positive regulator values for the LSTM-based hybrid anomaly detection models. While the thresholds for both Gaussian and Tukey's method based models vary significantly with each data set considered, the probability values for EVT-based detection is found to remain within [$10^{-3}$,  $10^{-5}$].}
\label{lstm-based}
\end{table*}

\subsubsection{EVT-LSTM Model}
The hyper-parameters and false positive regulators chosen for hybrid LSTM models are used for the EVT-LSTM model as well. We follow the guidelines in \cite{ruff2018deep} while setting the hyper-parameter $\lambda$ for the network weight regularizer. The threshold is updated every $k$ = 20 epochs. The values chosen for hybrid deep learning models seem to suit end-to-end deep learning models, for most of the scenarios considered. An exception was the Bengaluru Taxi Demand data set, where the suitable value for $q$ turned out to be $10^{-5}$. Nevertheless, the best choices for the probability $q$ remained in [$10^{-3}$, $10^{-5}$]. 
\begin{table}[h]
\begin{center}
{\renewcommand{\arraystretch}{1.1}
\begin{tabular}{ l c  c }
\hline
\multicolumn{2}{ l  }{\textbf{Data Sets}}                                                         & \textbf{P-values} \\ \hline \hline
\multicolumn{2}{ l  }{Vehicular Travel Time}                                            & 0.005    \\ \hline
\multicolumn{2}{ l  }{Vehicular Speed}                                                  & 0.005    \\ \hline
\multicolumn{2}{ l  }{Vehicle Occupancy}                                                & 0.370     \\ \hline
\multirow{3}{*}{\begin{tabular}[c]{@{}l@{}}NYC Taxi \\ Demand\end{tabular}}       & S1 & 0.805    \\ 
                                                                                  & S2 & 0.056    \\ 
                                                                                  & S3 & 0.147    \\ \hline
\multirow{3}{*}{\begin{tabular}[c]{@{}l@{}}Bengaluru Taxi \\ Demand\end{tabular}} & S1 & 0.570     \\ 
                                                                                  & S2 & 0.180     \\ 
                                                                                  & S3 & 0.006    \\ \hline
\multicolumn{2}{ l  }{Electrocardiogram}                                                & 0.002    \\ \hline
\multicolumn{2}{ l  }{Bitcoin Prices}                                                   & 0.051    \\ \hline
\end{tabular}}
\end{center}
\caption{P-values obtained from the A-D statistical test. The decision to reject the null hypothesis is taken when the p-values lie below 0.001. In all the data sets considered, the null hypothesis that the tails of the prediction errors follow a GPD is accepted.}
\label{pval}
\end{table}

% \usepackage{multirow}
\begin{table*}[]
\begin{center}
\scalebox{1}{
{\renewcommand{\arraystretch}{1.2}
\begin{tabular}{lccccccc}
\hline
\multicolumn{2}{l}{\multirow{4}{*}{\textbf{Data Sets}}}                                          & \multicolumn{6}{c}{\textbf{Anomaly Detection Models}}                                                                                                                                                                                                                                                        \\\cline{3-8}
\multicolumn{2}{l}{}                                                                   & GARCH        & OC-SVM & \begin{tabular}[c]{@{}c@{}}LSTM \\ Tukey \\ (Hybrid)\end{tabular} & \begin{tabular}[c]{@{}c@{}}LSTM\\ Gaussian \\ (Hybrid)\end{tabular} & \begin{tabular}[c]{@{}c@{}}LSTM \\ EVT \\ (Hybrid)\end{tabular} & \begin{tabular}[c]{@{}c@{}}EVT-LSTM\\ (End-to-End)\end{tabular} \\ \hline \hline
\multicolumn{2}{l}{\begin{tabular}[c]{@{}l@{}}Vehicular Travel Time\end{tabular}}   & 0.01         & 0.04   & 0.07                                                              & 0.21                                                                & \textbf{0.36}                                                   & \textbf{0.36}                                                   \\ \hline
\multicolumn{2}{l}{Vehicular Speed}                                                    & 0.18         & 0.56   & \textbf{0.79}                                                              & {0.74}                                                       & \textbf{0.79}                                                   & \textbf{0.79}                                                   \\\hline
\multicolumn{2}{l}{Vehicle Occupancy}                                                  & \textbf{1.0} & 0.33   & 0.5                                                               & \textbf{1.0}                                                        & \textbf{1.0}                                                    & \textbf{1.0}                                                    \\\hline
\multirow{3}{*}{\begin{tabular}[c]{@{}l@{}}NYC Taxi \\ Demand\end{tabular}}       & S1 & 0.002        & 0.03   & 0.25                                                              & \textbf{1.0}                                                        & \textbf{1.0}                                                    & \textbf{1.0}                                                    \\
                                                                                  & S2 & 0.005        & 0.16   & 0.14                                                              & 0.33                                                                & \textbf{1.0}                                                    & \textbf{1.0}                                                    \\
                                                                                  & S3 & 0.007        & 0.6    & 0.66                                                              & \textbf{0.86}                                                       & \textbf{0.86}                                                   & \textbf{0.86}                                                   \\\hline
\multirow{3}{*}{\begin{tabular}[c]{@{}l@{}}Bengaluru \\ Taxi Demand\end{tabular}} & S1 & 0.03         & 0.29   & 0.47                                                              & 0.57                                                                & \textbf{1.0}                                                    & \textbf{1.0}                                                    \\
                                                                                  & S2 & 0.002        & 0.12   & 0.08                                                              & 0.5                                                                 & 0.5                                                             & \textbf{0.66}                                                   \\
                                                                                  & S3 & 0.04         & 0.44   & 0.26                                                              & 0.54                                                                & 0.62                                                            & \textbf{0.72}                                                   \\\hline
\multicolumn{2}{l}{Electrocardiogram}                                                  & 0.1          & 0.22   & \textbf{0.49}                                                     & 0.32                                                                & 0.37                                                            &  0.28                                                               \\\hline
\multicolumn{2}{l}{Bitcoin Prices}                                                     & 0.52         & 0.31   & 0.19                                                              & 0.83                                                                & 0.83                                                            & \textbf{0.84}   \\ \hline                                              
\end{tabular}}}
\caption{The anomaly detection performance of various models considered in the study, across diverse data sets, based on F1-score. The proposed end-to-end EVT-LSTM deep anomaly detection model is observed to perform better compared to the statistical, machine learning and hybrid deep learning techniques considered.}
\label{results}
\end{center}
\end{table*}

\section{Results} \label{resultssection}
In this section, we analyze whether the tails of the prediction error distribution follow a GPD, and present results from the numerical tests performed.  

\subsection{Statistical Tests} \label{statistical}
We conduct a statistical test known as the A-D (Anderson-Darling) test \cite{stephens1974edf} for checking the compliance of the tail distribution to a GPD. The A-D test can be used to assess whether a sample of the data comes from a specific probability distribution. This test makes use of the specific distribution while calculating the critical values. The test statistic $A^2$ measures the distance between the hypothesized distribution and the empirical CDF of the data. Based on the test static and the p-values obtained, the null hypothesis that the data follow a specified distribution can (cannot) be rejected. The A-D test is a modification of the K-S (Kolmogorov-Smirnov) test \cite{massey1951kolmogorov} and gives more weight to the tails than does the K-S test. The A-D test is conducted on the excesses $X-T$, \emph{i.e.}, the prediction errors lying above empirical threshold $T$. The p-values obtained from this statistical test are given in Table \ref{pval}. We reject the null hypothesis for each data set if the corresponding p-value lies below 0.001. For all the data sets under study, statistical evidence from the A-D test suggests that the tail distributions of the prediction errors tend to follow GPD.

\subsection{Numerical Results}
The anomaly detection performance based on the F1-score metric, of various models across different data sets, is provided in Table~\ref{results}. Based on the results from the table, we can draw the following inferences:
\begin{itemize}
    \item The poor performance of the parametric GARCH models suggest that assuming a particular distribution on the prediction errors can critically affect anomaly detection accuracy.
    \item Deep learning-based anomaly detection algorithms exhibit superior detection accuracy over statistical and machine learning-based algorithms across seven diverse data sets.
    \item Out of the two classes of deep learning-based anomaly detection models considered, an end-to-end detection algorithm outperforms hybrid detection models on a broad variety of data sets.
\end{itemize}

When the parametric GARCH model is employed for anomaly detection, we observe that the model has a sufficiently high Recall, but very low Precision. The threshold chosen based on the validation set classifies a large number of non-anomalies as anomalous on the test set. Thus, the overall anomaly detection performance is affected by the presence of several false positives, resulting in a low F1-score value. Exceptions to this behavior are observed with vehicle occupancy data set and to an extent, with the bitcoin prices data. The magnitude of the anomalies is much higher than that of the non-anomalies in these data sets, which appears to be the reason behind this exception.

The OC-SVM model achieves a higher detection accuracy compared to statistical GARCH model but does not fare well compared to the deep learning variants. They also showcase high Recall and poor Precision values. On the other hand, a single value of kernel coefficient $\alpha$ (0.0001) proved to be a satisfactory fit for all the data sets considered.

On comparing hybrid and end-to-end deep anomaly detection models, we see that the proposed end-to-end EVT-LSTM model shows superior detection accuracy. The anomaly detection requires no post-processing tools, and the performance is always at least as good as that of the hybrid models considered, for the majority of data sets considered. This observation suggests that a deep learning model customized for anomaly detection can provide better accuracy results than running traditional algorithms on a deep learning model developed for forecasting. The only exception is observed in the ECG data set, which can be attributed to the anomaly labeling scheme followed. The labeling scheme employed in this data set marks an entire period of the ECG signal as anomalous in case any point in that period is an anomaly. In other words, we deal with \emph{collective} anomalies in this data set. The fraction of anomalies is, hence, higher in the ECG data set compared to other data sets that have point anomalies. Thus, the anomalies cover a broad spectrum above the upper quartile of prediction errors for ECG data. Since the Tukey's method thresholds the raw prediction errors based on the upper quartile, it results in good anomaly detection for the ECG data set. This finding suggests that a simple threshold based on the magnitude of prediction errors might be sufficient when the fraction of anomalies in the data set is relatively high. Generally, Tukey's method can detect most of the anomalies but results in a large number of false positives, similar to GARCH and OC-SVM models. This behavior is not desirable in an anomaly detection setting.  

An important observation is made regarding the variability in false positive regulator values of various methods. Recalling the results from Table \ref{lstm-based}, we find high variability in the false positive regulator values of Gaussian and Tukey detection rules. The choices for thresholds $\tau_g$ and $\tau_e$ vary significantly with the data set considered. While $\tau_g$ varied between [-15, -25], $\tau_t$ was found to take values between [0.11, 12961.8]. The strong dependence of the anomaly thresholds on the time-sequence considered limit the applicability of such detection rules. On the other hand, the only free parameter for EVT-based detection, the probability $q$, does not appear to have a significant dependence on the data set. This false positive regulator was found to stay within the range [$10^{-3}$, $10^{-5}$]. A false positive parameter with low dependency on the data sets is highly preferred in real-world settings, thereby strengthening the case of a detection algorithm based on EVT.

In summary, considering data sets from various verticals of ITS, we found that an end-to-end deep learning-based anomaly detection algorithm holds great potential in detecting abnormal traffic instances. Our proposed EVT-LSTM model accurately detected anomalous traffic speed, vehicle occupancy, travel time, and taxi demand instances, in addition to data sets from medical and financial domains. 

\section{Concluding Remarks}\label{conclusionsection}
Detection of anomalies is a crucial part of ITS (Intelligent Transportation Systems), as it can provide useful recommendations to urban planners and taxi aggregators, among others. In this study, we developed an end-to-end deep learning-based anomaly detection model for temporal data in transportation networks. 

The proposed EVT-LSTM model incorporates concepts from EVT (Extreme Value Theory) into the objective function of an LSTM (Long Short-Term Memory) deep learning model. The output network representations from our proposed model can be directly utilized for anomaly detection, a clear advantage over the currently popular hybrid deep learning-based detection models that require separate post-processing tools. 

Our proposed model was compared against traditional statistical, machine learning, and deep learning-based anomaly detection models. When evaluated across seven diverse data sets, the EVT-LSTM model exhibited superior anomaly detection performance against these established baseline models. The proposed model was able to detect true positives faithfully while incurring as few false positives as possible. We found strong evidence to suggest that a deep learning model customized for anomaly detection can provide better detection accuracy than the hybrid deep anomaly detection techniques. 

There are numerous avenues that merit future attention. To validate the performance of the proposed algorithm further, new data sets can be introduced. While our algorithm employs an objective function based on EVT, it would be useful to explore other objective functions, to enhance the detection accuracy.
% \balance
% \section*{Acknowledgements}

%\bibliographystyle{IEEEtran}

\end{document}